\documentclass[runningheads]{llncs}

\usepackage{graphicx}
\usepackage{comment}
\usepackage{hyperref}
\usepackage{macros}
\usepackage{algorithm} 
\usepackage{algpseudocode}
\usepackage{xcolor}
\usepackage{soul}

\usepackage{amsmath}
\usepackage{placeins}
\usepackage[inline]{enumitem}
\usepackage[all]{nowidow}

\newcommand{\corRH}[2]{{\color{cyan}#2}}

\begin{document}

\title{Open Set  Domain Adaptation using Optimal Transport}

\author{Marwa \textsc{Kechaou}, Romain \textsc{Herault}, Mokhtar Z. \textsc{Alaya} and Gilles \textsc{Gasso}}
\institute{Normandie Univ, UNIROUEN, UNIHAVRE, INSA Rouen, LITIS, 76 000 Rouen, France}

\newif\ifbibfile
\bibfiletrue 

\maketitle              
\begin{abstract}
We present a 2-step optimal transport approach that performs a mapping from a source distribution to a target distribution.
Here, the target has the particularity to present new classes not present in the source domain.
The first step of the approach aims at rejecting the samples issued from these new classes using an optimal transport plan.
The second step solves the target (class ratio) shift  still as an optimal transport problem.
We develop a dual approach to solve the optimization problem involved at each step and we prove that our results outperform recent state-of-the-art performances.
We further apply the approach to the setting where the source and target distributions present both a label-shift and an increasing covariate (features) shift to show its robustness.

\end{abstract}
\section{Introduction}
Optimal Transport (OT) approaches tackle the problem of finding an optimal mapping between two distributions $P^s$ and $P^t$ respectively from a source domain and a target domain by minimizing the cost of moving probability mass between them.
Efficient algorithms are readily available to solve the OT problem \cite{peyre2019COTnowpublisher}. 
 
A wide variety of OT applications has emerged ranging from computer vision tasks \cite{bonneel2011displacement} to machine learning applications \cite{courty2016optimal,pmlr-v70-arjovsky17a}.
Among the latter, a body of research work was carried out to apply OT to domain adaptation task \cite{courty2016optimal,courty2017joint,redko2017theoretical,wu2019domain}. 
Domain Adaptation (DA)  assumes  labelled samples $(x, y)$ in the source domain while only unlabelled (or a few labelled) data  are available in the target domain.
It intends to learn a mapping so that the prediction model tuned for the source domain applies to the target one in the presence of shift between source and target distributions. 
The distribution shift may be either a \textit{Covariate-Shift} where the marginal probability distributions $P^s(X)$ and $P^t(X)$ vary across domains while conditional probability distributions is invariant (i.e. $P^s(Y|X) = P^t(Y|X)$) or a \textit{Label-Shift} where label distributions $P(Y)$ for both domains do not match but their conditional probability distributions $P(X|Y)$ are the same.
Theoretical works \cite{Ben-David2010,pmlr-v97-zhao19a} have investigated the generalization guarantees on target domain when transferring knowledge from the labeled source data to the target domain. 

Courty et al. \cite{courty2016optimal} settled OT to deal with covariate-shift by enforcing samples from a class in the source domain to match with the same subset of samples in the target domain.
Follow up works extend OT to asymmetrically-relaxed matching between the distributions $P^s(X)$ and $P^t(X)$ or to joint distribution $P(X,Y)$ matching between source and target domains \cite{courty2017joint,bhushan2018deepjdot}.
Recently, Redko et al. \cite{ievgen_2019} focus on multi-source domain adaptation under target shift and aim to estimate the proper label proportions $P^t(Y)$ of the unlabelled target data.
Traditional DA methods for classification  commonly assume that the source and target domains share the same label set.
However in some applications, some source labels may not appear in the target domain.
This turns to be an extreme case of label-shift when the related target class proportions drop to zero.
The converse case, termed as open set domain adaptation \cite{panareda2017open}, considers a target domain with additional labels which are deemed abnormal as they are \emph{unknown classes} from the source domain standpoint.
This results in  a substantial alteration in the label distributions as  $P^s(y_k)=0$ and $P^t(y_k) \neq 0$ for some labels $y_k$ not occurring in the source domain.
Therefore, aligning the label distributions $P^s(Y)$ and $P^t(Y)$ may lead to a negative transfer.
To tackle this issue, open set domain adaptation aims at rejecting the target domain ``abnormal samples" while matching the samples from the shared categories \cite{panareda2017open,saito2018open,fang2019open}.

In this paper, we address the open set DA using optimal transport.
The approach we propose consists of the following two steps: 
\begin{enumerate*}[label=\arabic*)]
\item rejection of the outlier samples from the unknown classes  followed by
\item a label shift adaptation.
\end{enumerate*}
Specifically, we frame the rejection problem as learning an optimal transport map together with the target marginal distribution $P^t(X)$ in order to prevent source samples from sending probability mass to unknown target samples.
After having rejected the outliers from target domain, we are left with a label shift OT-based DA formulation. Contrary to the first step, we fix the resulting target marginal $P^t(X)$ (either to a uniform distribution or to the $P^t(X)$ learned at the first stage) and optimize for a new transport map and the source marginal distribution $P^s(X)$ in order to re-weight source samples according to the shift in the proportions of the shared labels. 
We also propose a decomposition of $P^s(X)$ and show its advantage to reduce the number of involved parameters. 
To the best of our knowledge, this is  the first work considering open set DA problem using OT approach. The key contributions of the paper are:
\begin{enumerate*}[label=\roman*)]
\item We devise an OT formulation to reject samples of unknown class labels by simultaneously optimizing  the transport map and the target marginal distribution.
\item We propose an approach to address the label-shift which estimates the target class proportions and enables the prediction of the target sample labels.

\item We develop the dual problem of each step (rejection and label-shift) and give practical algorithms to solve the related optimization problems.
\item We conduct several experiments on synthetic and real-datasets  to assess the effectiveness of the overall method.
\end{enumerate*}
 
The paper is organized as follows: in Section~\ref{sec:related_work} we detail the related work. Section~\ref{sec:our_approach} presents an overview of discrete OT, our approach, and the dual problem of each step. It further details the optimization algorithms  and some implementation remarks.
Section~\ref{sec:expes} describes the experimental evaluations.

\section{Related Work} \label{sec:related_work}

Arguably the most studied scenario in domain adaptation  copes with the change in the marginal probability distributions $P^s(X)$ and $P^t(X)$. 
 
Only a few dedicated works have considered the shift in the class distributions $P^s(Y)$ and $P^t(Y)$. 

To  account for the label-shift, Zhang et al. \cite{zhang2013domain} proposed a re-weighting scheme of the source samples.
The weights are determined by solving a maximum mean matching problem involving  the kernel mean embedding of the marginal distribution $P^s(X)$ and the conditional one $P^s(X/Y)$.
In the same vein, Lipton et al. \cite{LiptonWS18}  estimated  the weights $P^t(y_k)/P^s(y_k)$ for any label $y_k$ using a black box classifier elaborated on the source samples.
The estimation relies on the confusion matrix and on the approximated target class proportions via the pseudo-labels given by the classifier.

The re-weighting strategy is also investigated in the JCPOT procedure \cite{ievgen_2019} using OT and under multiple source DA setting.
The target class proportions are computed by solving a constrained Wassesrtein barycenter problem defined over the sources. Wu et al. \cite{wu2019domain} designed  DA with asymmetrically relaxed distribution alignment to lift the adversarial DA approach \cite{ganin2016domain} to label-shift setup. 
Of a particular note is the label distribution computation \cite{sanderson2014class} which hinges on mixture proportion estimation. The obtained class proportions can be leveraged on to adapt source domain classifier to target samples.
Finally, JDOT approach \cite{courty2017joint} addresses both the covariate and label shifts by aligning  the joint distributions $P^s(X, Y)$ and $P^t(X, \hat Y)$ using OT.
As in \cite{LiptonWS18} the target predictions $\hat Y$ are given by a classifier learned jointly with the related OT map.

Regarding open set DA, the underlying principle of the main approaches resembles the one of multi-class open set recognition or multi-class anomaly rejection (see  \cite{sanderson2014class,article,10.1007/978-3-319-10578-9_26,6809169} and references therein) where one looks for a classifier with a reject option. 
\cite{panareda2017open} proposed an iterative procedure combining assignment and linear transformation to solve the open set DA problem.
The assignment step consists of a constrained binary linear programming which ensures that any target sample is either assigned to a  known source class (with some cost based on the distance of the target sample to the class center) or labelled as outlier.
Once the unknown class samples are rejected, the remaining target data are matched with the source ones using a linear mapping.
Saito et al. \cite{saito2018open} devised an adversarial strategy where a generator is trained to indicate whether a target sample should be discarded or matched with the source domain.
Recently, Fang et al. \cite{fang2019open} proposed a generalization bound for open set DA and thereon derived a so-called distribution alignment with open difference in order to sort out the unknown and known target samples.
The method turns to be a regularized empirical risk minimization problem. 

\section{The proposed approach} \label{sec:our_approach}

We assume the existence of a labeled source dataset $Z^s = \{(x_1^s,y^s_1),..,(x_{n_s}^s,y^s_{n_s})\}$ where $\{y_i\}^{n_s}_{i=1} \in \{1,..,C\}^{n_s}$ with $C$ the number of classes and $n_s$ the number of source samples. We also assume available a set of $n_t$ unlabeled target samples  $Z^t = \{x_1^t,\ldots,x_{n_t}^t\}$. {The target samples are assumed to be of labels $y^t$ in $\{1,..,C, C+1\}$ where the class $C+1$ encompasses all target samples from other classes not occurring in the source domain. Moreover we assume that the proportions of the shared classes may differ between source and target domains i.e. $P^s(y) \neq P^t(y)$ for $y\in \{1,..,C\}$. }

{Let $P^s(x|y)$ and $P^t(x|y)$ be the conditional distributions of source and target respectively with possibly $P^s(x|y) \neq P^t(x|y)$ for $y\in \{1,..,C\}$. Similarly we denote the marginal source and target distributions as $P^s(x)$ and $P^t(x)$. 
Our goal is to learn a distribution alignment scheme able to reject from $Z^t$ the samples of the unknown class $C+1$  while matching correctly the remaining source and target samples by accounting for the label-shift and possibly a shift in the conditional distributions. For this, we propose a two-step approach (see in Fig.~\ref{fig: full algo} in the appendix~\ref{ap: full algo} for an illustration): a rejection step followed by the label shift correction. To proceed we rely on discrete OT framework which is introduced hereafter.}

\subsection{The general optimal transport framework}

This section reviews the basic notions of discrete OT and fixes additional notation.
Let $\Sigma_{n}$ be the probability simplex with $n$ bins, namely the set of probability vectors in $\mathbb{R}_{+}^{n},$ i.e., $\Sigma_{n}=\big\{\omega\in \mathbb{R}^{n}_+: \norm{\omega}_1:= \sum_{j=1}^{n} \omega_{j}=1\big\}$.
Let $\mu^s$ and $\mu^t$ be  two discrete distributions derived respectively from $\Sigma_{n_s}$ and $\Sigma_{n_t}$ 
such that 
\begin{equation}
\mu^s = \sumins \mu^s_i \delta_{x^s_i} \text{ and } \mu^t = \sumint \mu^t_j \delta_{x^t_j},  \nonumber
\end{equation}
where $\mu^s_i$ stands for the probability mass associated to the $i$-th sample (the same for $\mu^t_j$). Computing OT distance between $\mu^s$ and $\mu^t$, referred to as the Monge-Kantorovich or Wasserstein distance~\cite{kantorovich1942,villani03topics}.
amounts to solving the linear problem given by 
\begin{equation}
\label{wasserstein-dist}
W(\mu^s, \mu^t) = \min_{\gamma \in \Pi(\mu^s, \mu^t)} \inr{\zeta, \gamma}_F,
\end{equation}
where $\inr{\cdot, \cdot}_F$ denotes Frobenius product between two matrices, that is
$\inr{T,W}_F = \sum_{i,j}T_{ij}W_{ij}$. Here the matrix $\zeta = (\zeta_{ij})_{1\leq i \leq n_s; 1\leq j \leq n_t} \in \R^{n_s \times n_t}_+$, where  each $\zeta_{ij}$ represents the energy needed to move a probability mass from $x^s_i$ to $x^t_i$. In our setting $\zeta$ is given by the pairwise Euclidean distances between the instances in the source and target distributions, i.e., $\zeta_{ij} = \norm{x_i^s - x_j^t}_2$.
The matrix $\gamma= (\gamma_{ij}) \in \R^{n_s\times n_t}_+$ is called a transportation plan, namely each entry $\gamma_{ij}$ represents the fraction of mass moving from $x_i^s$ to $x_j^t$. The minimum $\gamma$'s in problem~\eqref{wasserstein-dist} is taken over the convex set of probability couplings  between $\mu^s$ and $\mu^t$ defined by 
\begin{equation*}
\Pi(\mu^s, \mu^t) = \{\gamma \in \R^{n_s \times n_t}_+: \gamma \mathbf 1_{n_t} = \mu^s, \gamma^\top \mathbf 1_{n_s} = \mu^t\},
\end{equation*}
where we identify the distributions with their probability mass vectors, i.e. $\mu^s \equiv (\mu^s_1, \ldots, \mu^s_{n_s})^\top$ (similarly for $\mu^t)$, and 
$\mathbf 1_n \in \R^n$ stands for all-ones vector. 
The set $\Pi(\mu^s, \mu^t)$ contains all possible joint probabilities with marginals corresponding to $\mu^s$ and $\mu^t.$
In the sequel 
when applied to matrices and vectors, product, division and exponential notations refer to element-wise operators.

Computing classical Wassertein distance is computationally expensive, since its Kantorovich fomulation~\eqref{wasserstein-dist} is a standard linear program with a complexity $O(\max(n_s, n_t)^3)$ \cite{leeSidford2013PathFI}. To overcome this issue, a prevalent approach, referred to as regularized OT~\cite{cuturinips13}, operates by adding an entropic regularization penalty to the original problem and it writes as 
\begin{equation}
\label{wasserstein-dist-entrop}
W_\eta(\mu^s, \mu^t) = \min_{\gamma \in \Pi(\mu^s, \mu^t)} \{\inr{\zeta, \gamma}_F - \eta H(\gamma)\}
\end{equation}
where $H(\gamma) = -\sumins\sumint \gamma_{ij}\log \gamma_{ij}$ defines the entropy of the matrix $\gamma$ and $\eta>0$ is a  regularization parameter to be chosen. Adding the entropic term makes the problem significantly more amenable to computations. In particular, it allows to solve efficiently the optimization problem~\eqref{wasserstein-dist-entrop} using a balancing algorithm known as Sinkhorn's algorithm~\cite{sinkhorn1967}. Note that the Sinkhorn iterations are based on the dual solution of~\eqref{wasserstein-dist-entrop} (see~\cite{peyre2019COTnowpublisher} for more details).

\subsection{First step: Rejection of unknown class samples}

In the open set DA setting, a naive application of the preceding OT framework to source set $Z^s = \{(x_1^s,y^s_1),..,(x_{n_s}^s,y^s_{n_s})\}$ and target dataset  $Z^t = \{x_1^t,\ldots,x_{n_t}^t\}$ will lead to undesirable mappings as some source samples will be transported onto the abnormal target samples. To avoid this, we intend to learn a transportation map such that the probability mass sent to the  unknown abnormal samples of the target domain will be negligible, hence discarding those samples. A way to achieve this goal is to adapt the target marginal distribution $P^t(X)$ while learning the map.  

Therefore, to discard the new classes appearing in the target domain, in a first stage, we {solve} the {following optimization} problem:
\begin{equation}
\gamma^\star_{\rj}, \mu^{t^\star}  = \argmin_{\substack{\gamma \in \Pi(\mu^s, \mu^t)\\ \mu^t \in \Sigma_{n_t}}}\{\inr{\zeta, \gamma}_F - \eta H(\gamma)\},
\label{rejection-prb}
\end{equation}
where $\mu^t$ stands for the target marginal $P^t(X)$ and $\mu^s$ for the source one $P^s(X)$. 

{The first stage of the rejection step as formulated in~\eqref{rejection-prb} aims at calculating a transportation plan while optimizing the target marginal $\mu^t$.
The rationale for updating $\mu^t$  
is linked to the new classes appearing in target domain.
Therefore, the formulation allows some freedom on $\mu^t$ and leads to more accurate matching between known marginal source and unknown marginal target.
}
{To solve this optimization problem}, we use Sinkhorn iterations~\cite{cuturinips13}. Towards this end, we explicit its dual form in Lemma~\ref{lemma_rejection}. Hereafter, we set $B(f, g) = \text{diag}(e^{f}) K \text{diag}(e^{g})$ where $K = e^{-\zeta/\eta}$ stands for the Gibbs kernel associated to the cost matrix $\zeta$ and where $\text{diag}$ denotes the diagonal operator.

\begin{lemma}
\label{lemma_rejection}
The dual problem of~\eqref{rejection-prb} reads as 
\begin{align}
\label{dual-rejection-prb}
(f^\star_{\rj}, g^\star_{\rj}) 
= \argmin_{\substack{f \in \R^{n_s}, g \in \R^{n_t}}} \{\mathbf 1_{n_s}^\top B(f, g) \mathbf 1_{n_t} -\inr{f, \mu^s} + \chi_{-\mathbf 1_{n_t}} (g)\},
\end{align}
where for all $g \in \R^{n_t}$ we denote by 
\begin{equation*}
\chi_{-\mathbf 1_{n_t}} (g) = \begin{cases}
   0, &\text{ if } g = -\mathbf 1_{n_t}, i.e.\, g_j = -1, \forall j = 1, \ldots, n_t,\\
   \infty, & { otherwise}.
\end{cases}
\end{equation*}
\end{lemma}
Note that the optimal solutions $\gamma^\star_{\rj}$ and $\mu^{t^\star}$ of the primal problem take the form
\begin{equation*}
    \gamma^\star_{\rj} =B(f^\star_{\rj}, g^\star_{\rj}), \quad \mu^{t^\star} = \gamma_{\rj}^{\star \top} \mathbf 1_{n_s}.
\end{equation*}

{Once $\mu^{t}$ is learned, the second stage consists in discarding the new classes by relying on the values of $\mu^{t^\star}$.
{Specifically, we reject} the $j$-th sample in the target set whenever $\mu^{t^\star}_j$ is a neglectable value with respect to some chosen threshold\corRH{, see Section~\ref{sec_num_expes_rejection}}{}.
Indeed, since  $\gamma^\star_{\rj}$ satisfies the target marginal constraint $\mu^{t^\star}_j = \sumins (\gamma^\star_{\rj})_{ij}$ for all $j=1, \ldots, n_t$, we expect that the row entries $\{(\gamma^\star_{\rj})_{ij}: i=1, \ldots,n_s\}$ take small values for each $j$-th sample associated to a new class, that is we avoid transferring  probability mass from source samples to the unknown target $j$-th  instance.
{The tuning of the rejection threshold is exposed in  Section~\ref{sec_num_expes_rejection}.}
}

{The overall rejection procedure is  depicted in Algorithm~\ref{algo_rejection}.} To grasp the elements of Algorithm \ref{algo_rejection} and its stopping condition, we  derive the Karush-Kuhn-Tucker (KKT) optimality conditions~\cite{boyd2004convex} for the rejection dual problem (Eq.~\ref{dual-rejection-prb}) in Lemma~\ref{lem_kkt_rejection}.  

\begin{lemma}
\label{lem_kkt_rejection}
The couple $(f^\star_{\rj}, g^\star_{\rj})$ optimum of problem~\eqref{dual-rejection-prb} satisfies
\begin{equation}
\label{kkt_f_rj}
(f^\star_{\rj})_i = \log(\mu^s_i) - \log\Big(\sumint K_{ij}e^{(g^\star_{\rj})_j}\Big)
\end{equation}
and 
\begin{equation}
\label{kkt_g_rj}
 \sumint e^{(f^\star_{\rj})_i} K_{ij} e^{(g^\star_{\rj})_j} (1 + (g^\star_{\rj})_j) \leq 0,
\end{equation}
for all $i=1, \ldots, n_s.$
\end{lemma}


The proofs of Lemma~\ref{lemma_rejection} and~\ref{lem_kkt_rejection} are postponed to Appendix~\ref{ap:lemma_rejection}.

We remark that we have a closed form of $f^\star_{\rj}$, see Eq.~\ref{kkt_f_rj}, while it is not the case for $g^\star_{\rj}$ as shown in  Eq.~\ref{kkt_g_rj}.
This is due to non-differentiability of the objective function defining the couple $(f^\star_{\rj},  g^\star_{\rj})$. 
Therefore, we tailor Algorithm~\ref{algo_rejection}  with a sufficient optimality condition to guarantee Eq.~\ref{kkt_g_rj}, in particular we set $(g^\star_{\rj})_j \leq -1$ for all $j=1, \ldots, n_t$.
These latter conditions can be tested on the update of the target marginal $\mu^t$ for the rejection problem (see Steps 6-9 in Algorithm~\ref{algo_rejection}).
We use the condition $\norm{B(f,g) \mathbf 1_{n_t} - \mu^s}_1 + \norm{B(f,g)^\top \mathbf 1_{n_s} - \mu^t}_1 \leq \varepsilon$ ($\varepsilon$-tolerance) as a stopping criterion for Algorithm~\ref{algo_rejection}, which  is very natural since it requires that $B(f,g)\mathbf 1_{n_t} $ and $B(f,g)^\top \mathbf 1_{n_s}$ are close to the source and target marginals $\mu^s$ and $\mu^t.$

\begin{algorithm}[ht]
	\caption{Rejection (see Equation~\ref{rejection-prb})}
	\label{algo_rejection}
	\begin{small}
	\textbf{require}: $\mathbf{\eta}$: regularization parameter, $\mathbf{\zeta}$: cost matrix, \textbf{$Z^t$}: target samples, \textbf{$n_s$}: number of source samples, \textbf{$n_t$}: number of target samples, 
	\textit{tol}: tolerance, \textit{thresh}: threshold;\\
	\textbf{output}: transport matrix: $\gamma_{\rj} = B(f_{\rj}, g_{\rj})$; target marginal: $\mu^t;$ rejected samples: $X_{\rj}^t$
	\begin{algorithmic}[1]
		\State \textbf{initialize:}
		\State $err \gets 0;$  $f \gets \mathbf 0_{n_s};$
		$g \gets - \mathbf 1_{n_t};$ $\mu^t$ $\gets$ $\frac{1}{n_t} \mathbf 1_{n_t};$
		\While{$err > tol$}
		\State $f \gets \log(\mu^s) - \log(Ke^{g});$
		\State $\mu^t \gets B(f, g)^\top \mathbf 1_{n_s};$ 
		\ForAll {$j=1, \ldots ,n_t$}
		\If{$\mu^t_{j} > e^{-1} \sumins K_{ij} e^{g_j}$}
		\State $\mu^t_j \gets e^{-1} \sumins K_ij e^{g_j};$
		\EndIf
		\EndFor
		\State  $g \gets$ log($\mu^t$) - log($K^\top e^{f}$);
		\State $err \gets$ $\norm{B(f,g) \mathbf 1_{n_t} - \mu^s}_1 + \norm{B(f, g)^\top \mathbf 1_{n_s} - \mu^t}_1;$
		\EndWhile
		\State $Z_{\rj}^t \gets$ $Z^t[\mu^t \leq thresh]$
		\State \textbf{return:} $B(f,g)$, $\mu^t$ and $Z_{\rj}^t$
	\end{algorithmic}
	\end{small}
\end{algorithm} 

\subsection{Second step: Label-Shift correction}

We re-weight source samples to correct the difference in class proportions between source and target domains. Correcting the label shift is formulated as 
\begin{equation}
\label{label-shift-prb} 
\gamma^\star_{\ls}, \nu^\star = \argmin_{\substack{\gamma \in \Pi(D\nu, \mu^t)\\ \nu \in \Delta_C}}\{\inr{\zeta, \gamma}_F - \eta H(\gamma)\},
\end{equation}
where the target marginal $\mu^t$ is either a uniform distribution or the one learned at the rejection step and where $D = (d_{ic})\in \R^{n_s \times C}_+$ is a linear operator, such that for $i=1, \ldots, n_s$ and $c=1, \ldots, C$  
\begin{equation*}
d_{ic} = \begin{cases}
   \frac{1}{n_s^c}, &\text{ if } y_i^s = c, \\
   0, & \text{ otherwise}.
\end{cases}
\end{equation*}
Here $n_s^c$ denotes the cardinality of source samples with class $c$, namely $n_s^c = \#\{i=1, \ldots, n_s:y_i^s = c\}.$  The parameter vector $\nu = (\nu_c)_{c=1}^C$ belongs to the convex set
\begin{equation*}
\Delta_C = \Big\{\alpha \in \R^{C}_+:\sum_{c=1}^C\sumins d_{ic}\alpha_c = 1\Big\}.
\end{equation*}
{
In order to estimate the unknown class proportions in the target domain, we set-up the source marginal as $\mu^s = D\nu$ where the entry $\nu_c$ expresses the $c$-class proportion for all $c=1, \ldots, C.$ Once we estimate theses proportions, we can get the class proportions in target domain thanks to OT matching.
We shall stress that Problem (\ref{label-shift-prb}) involves the simultaneous calculation of the transportation plan $\gamma_{\ls}$ and the source class re-weighting. Our procedure resembles the re-weighting method of JCPOT \cite{ievgen_2019} except that we do not rely on a Wasserstein barycentric problem required by the multiple source setting addressed in \cite{ievgen_2019}. 
The estimation $\nu^\star$ can be explicitly calculated using the source marginal constraint satisfied by the transportation plan $ \gamma^\star_{\ls}$, i.e., 
\begin{equation*}
\nu^\star = (D^\top D)^{-1} D^\top \gamma^\star_{\ls}\mathbf 1_{n_t}.
\end{equation*}
As for the rejection step, we use Sinkhorn algorithm  with an update on the source marginal $\mu^s = D\nu$ to solve the label shift Problem~(\ref{label-shift-prb}) via its dual as stated in Lemma~\ref{lemma_shift}.
}
\begin{lemma}
\label{lemma_shift}
The dual of problem~(\ref{label-shift-prb}) writes as 
\begin{align}
\label{dual-shift-prb}
(f^\star_{\ls}, g^\star_{\ls}) 
= \argmin_{\substack{f \in \R^{n_s}, g \in \R^{n_t}}} \{\mathbf 1_{n_s}^\top B(f, g) \mathbf 1_{n_t} -\inr{g, \mu^t} + \chi_{\mathcal{F}}(f)\},
\end{align}
 where $\mathcal{F} = \big\{f \in \R^{n_s}: \sumins (f_i  + 1)d_{ic} = 0, \forall c =1, \ldots, C\big\}$
and 
\begin{equation*}
\chi_{\mathcal{F}} (f) = 
\begin{cases}
   0, &\text{ if } f \in \mathcal{F},\\
   \infty, & \text{ otherwise}.
\end{cases}
\end{equation*}
\end{lemma}

 Moreover, the closed form of the transportation plan in the Label-Shift step is given by 
\begin{equation*}
    \gamma^\star_{\ls}  = B(f^\star_{\ls}, g^\star_{\ls}).
\end{equation*}

The analysis details giving the dual formulation in Equation~(\ref{dual-shift-prb}) in Lemma~\ref{lemma_shift} are presented in the appendices.
{ As for the rejection problem, the optimality conditions of the Label-Shift problem are described in the dedicated Lemma~\ref{lem_kkt_shift}} which proof is differed to Appendix~\ref{ap:lemma_shift}.

\begin{lemma}
\label{lem_kkt_shift}
The couple $(f^\star_{\ls}, g^\star_{\ls})$ optimum of problem~\eqref{dual-shift-prb} satisfies
\begin{equation}
\label{kkt_g_ls}
(g^\star_{\ls})_j = \log(\mu^t_j) - \log\Big(\sumins K_{ij}e^{(f^\star_{\ls})_i}\Big)
\end{equation}
and 
\begin{equation}
\label{kkt_f_ls}
\sumins  e^{(f^\star_{\ls})_i} K_{ij} e^{(g^\star_{\ls})_j} (1 + (f^\star_{\ls})_i) \leq 0,
\end{equation}
for all $j=1, \ldots, n_t.$
\end{lemma}

Algorithm~\ref{algo_shift} shows the related optimization procedure. Similarly to the rejection problem, we see that  $g^\star_{\ls}$ admits a close form (Eq.~\ref{kkt_g_ls}), while   $f^\star_{\ls}$ does not. As previously,  we endow the Algorithm~\ref{algo_shift} with the sufficient optimality conditions  (\ref{kkt_f_ls}) by ensuring $(f^\star_{\ls})_i \leq -1$ for all $i=1, \ldots, n_s$.
The conditions are evaluated  on the source marginal $\mu^s = D\nu^\star$  (see Steps 5-9 in Algorithm~\ref{algo_shift}).
Finally we use the same $\varepsilon$-tolerance stopping condition $\norm{B(f,g) \mathbf 1_{n_t} - \mu^s}_1 + \norm{B(f,g)^\top \mathbf 1_{n_s} - \mu^t}_1 \leq \varepsilon$.

\begin{algorithm}[ht]
	\caption{Label-Shift (see Equation ~\ref{label-shift-prb})}
	\label{algo_shift}
	\begin{small}
	\textbf{require}: $\mathbf{\eta}$: regularization term; $\mathbf{\zeta}$: cost matrix, $Y^s$: source labels, $n_s$: number of source samples; $n_t$: number of target samples; $C$: number of classes; 
	\textit{tol}: tolerance;
	\textit{D}: linear operator;\\
	\textbf{output}: transport matrix: $\gamma_{\text{ls}} = B(f_{\text{ls}}, g_{\text{ls}})$; class proportions: $\nu$;  Prediction of target labels: $\hat Y^t$
	
	\begin{algorithmic}[1]
		\State \textbf{initialize:}
		$err \gets 1;$ 
		$\nu \gets \frac{1}{C} \mathbf 1_{C};$
		$f \gets -\mathbf 1_{n_s};$ $g \gets \mathbf 0_{n_t};$  $ A \gets (D^\top D)^{-1}D^\top;$
		\While{$err > tol$}
		\State $g \gets \log( \mu^t) - \log(K^\top e^{f});$
		\State $\mu^s \gets D \nu;$
		\ForAll {$i=1,\ldots,n_s$}
		\If{$\mu^s_{i} < e^{-1} \sumint K_{ij} e^{g_j}$}
		\State $\mu^s_i \gets e^{-1} \sumint K_{ij} e^{g_j};$
		\EndIf
		\EndFor
		\State $f \gets \log(\mu^s) - \log(Ke^{g})$;
		
		\State $\nu \gets AB(f, g)\mathbf 1_{n_t};$
		\State  $err \gets \norm{B(f,g)1_{n_t} - \mu^s}_1 + \norm{B(f,g)^\top \mathbf 1_{n_s} - \mu^t}_1;$
		\EndWhile
		\State $\hat Y^t \gets \argmax(D^{\top}B(f,g));$//indices of the max. values of $D^{\top}B(f,g)$'s columns~\cite{ievgen_2019}
		\State \textbf{return:} $B(f,g)$, $\nu$ and $\hat Y^t$
	\end{algorithmic}
	\end{small}
\end{algorithm} 


%
\subsection{Implementation details and integration}
\label{sec:kkt}

Our proposed approach to open set DA performs samples rejection followed by sample matching in order to predict the target labels (either outlier or known source domain label). 
{Hence, at the end of each step, we identify either rejected samples or predict target labels (see Step 14 of Algorithms ~\ref{algo_rejection} and~\ref{algo_shift}). 

For rejection, we compare the learned target marginal $\mu^{t^\star}$ to some threshold  to recognize the rejected samples (See  Section \ref{sec_num_expes_rejection} for its tuning).
To fix the threshold,  we assume that the target samples that receive insufficient amount of probability mass coming from source classes likely cannot be matched to any source sample and hence are deemed outliers. 

To predict the labels of the remaining target samples, we rely on the transportation map $\gamma_{\ls}^\star$ given by Algorithm~\ref{algo_shift}. Indeed for Label-Shift, JCPOT \cite{ievgen_2019} suggested a label propagation approach to estimate labels from $N$ transportation maps (corresponding to $N$ source domains). 
Following JCPOT, 
the obtained transport matrix $\gamma_{\ls}^\star$ 
is proportional to the target class proportions.
Therefore, we estimate the labels of the target samples based on the probability mass they received from each source class using  $\hat Y^t  = \argmax (D^{\top} \, \gamma_{\ls}^\star)$.
The term $D^{\top} \, \gamma_{\ls}^\star$ provides a matrix of mass distribution over classes.}

Finally, we stress that the rejection and Label-Shift steps are separately done allowing us to compare theses approaches with the sate-of-art. Nevertheless, we can make a joint 2-step, that means after rejecting the instances with new classes in the target domain we plug the obtained target marginal $\mu^{t^\star}$ in the Label-Shift step. 
{Experimental evaluations show that similar performances are attained for separate and joint steps.}

\section{Numerical experiments} \label{sec:expes}

To assess the performance of each step, we first present the evaluations of Rejection and Label-Shift algorithms so that we can compare them to state-of-the-art approaches. Then we present overall accuracy of the joint 2-step algorithm.

\subsection{Abnormal sample rejection}
\label{sec_num_expes_rejection}

We frame the problem as a binary classification where common and rejected classes refer respectively to positive and negative classes.
Therefore, source domain has only one class (the positive) while target domain includes a mixture of positives and negatives.
We estimate their proportions and compare our results to open set recognition algorithms for unknown classes detection.

To reject the target samples, we lay on the assumption that they correspond to entries with a small value in $\mu^{t^\star}$. The applied threshold to these entries is strongly linked to the regularization parameter $\eta$ of the OT problem (\ref{rejection-prb}).
We remark that when $\eta$ increases, the threshold is high and vice versa, making the threshold proportional to $\eta$. Also {experimentally, we notice that the threshold has the same order of magnitude of $1/(n_s+n_t)$. Therefore, we define a new hyper-parameter $\alpha$ such that the desired treshold is given by  $\lambda = \alpha \frac{\eta}{n_s+n_t}$.}

In order to fix the hyper-parameters ($\eta$,$\alpha$) of the Rejection algorithm, we resort to Reverse Validation procedure \cite{4803844,10.1007/978-3-642-15939-8_35}.
For a standard classification problem where labels are assumed to be only available for source samples, a classifier is trained on $\{X^s,Y^s\}$ in the forward pass and  evaluated on $X^t$ to predict $\hat Y^t$.
In the backward pass, the target samples with the pseudo-labels  $\{X^t,\hat Y^t\}$ are used to retrain the classifier with the same hyper-parameters used during the first training, to predict $\hat Y^s$.
The retained hyper-parameters are the ones that provide  the best accuracy computed  from $\{Y^s,\hat Y^s\}$ without requiring  $Y^t$.

We adapt the reverse validation principle to our case. For fixed $(\eta, \alpha)$, Algorithm~\ref{algo_rejection} is run to get $\mu^t$ and to identify abnormal target samples. These samples are removed from $X^t$ leading to $X^t_{\rj}$. Then the roles of $X^s$ and $X^t_{\rj}$ are reversed. 
By running the Rejection algorithm to map $X^t_{\rj}$ onto $X^s$ we expect that the yielded marginal $\mu^s$ will have entries greater than the threshold $\lambda$.  
This suggests that we did not reject erroneously the target samples during the forward pass. 
As we may encounter mis-rejection, we select the convenient hyper-parameters ($\eta$,$\alpha$) 
that correspond to the highest $\frac{\#(\mu^t \leq \lambda)}{n_s}$. Algorithm~\ref{algo_RV} in Appendix~\ref{ap:reverse validation} gives the implementation details of the adapted Reverse Validation approach.

We use a grid search to find optimal hyperparameters ($\eta$,$\alpha$). $\eta$ was searched in the following set \{0.001,0.01,0.05,0.1,0.5,1,5,10\} and $\alpha$ in \{0.1,1,10\}. We apply Algorithm~\ref{algo_RV} and get $\eta =0.1$ and $\alpha=1$ for synthetic data and $\eta =0.01$ and $\alpha=10$ for real datasets.
\FloatBarrier

\subsubsection{Experiments on synthetic datasets}

We use a mixture of 2D Gaussian dataset with $3$ classes.
We choose $1$ or $2$ classes to be rejected in target domain as shown in Table~\ref{table:f1synth1}.  We generate $1000$ samples for each class in both domains with varying noise levels. 

The change of rejected classes  at each run induces a  distribution shift between shared (Sh) and rejected (Rj) class proportions.
Tables~\ref{table:f1synth1} and \ref{table:f1synth5} present the recorded F1-score. For a fair comparison, we tune the hyper-parameters of the competitor algorithms and choose the best F1-score for each experiment.

\begin{table}[ht]
\setlength{\tabcolsep}{4pt}
\begin{center}
\caption{F1-score on target domain for the  Rejection algorithm applied to synthetic dataset, Noise level $=$ 0.5 and $\eta$ $=$ 0.1}
\label{table:f1synth1} 
\begin{tabular}{l|llllll}
\hline\noalign{\smallskip}
Sh classes & \{0,1\} & \{0,2\} & \{1,2\} & \{0\} & \{1\} & \{2\}\\
Rj classes & \{2\}& \{1\} & \{0\} & \{1,2\} & \{0,2\} & \{0,1\}\\
\hline\noalign{\smallskip}
\% of Rj classes& 33\% & 33\% & 33\% & 66\% & 66\% & 66\%\\
\hline\noalign{\smallskip}
1Vs (Linear) & 0.46 & 0.5 & 0 & 0 & 0 & 0\\
\hline\noalign{\smallskip}
WSVM (RBF) & 0.99 & \textbf{0.99} & 0.99 & - & - & -\\
\hline\noalign{\smallskip}
PISVM (RBF) & 0.99 & \textbf{0.99} & 0.69 & 0.5 & 0.5 & 0.5\\
\hline\noalign{\smallskip}
Ours &  \textbf{1} & \textbf{0.99} & \textbf{0.99} & \textbf{1} & \textbf{1} & \textbf{0.98}\\
\hline
\end{tabular}
\end{center}
\setlength{\tabcolsep}{1.4pt}
\vspace*{-10mm}
\end{table}

\begin{table}[ht]
\setlength{\tabcolsep}{4pt}
\begin{center}
\caption{F1-score on target domain for the Rejection algorithm applied to synthetic dataset, Noise level $= 0.75$  and $\eta$ $=$ 0.5}
\label{table:f1synth5} 
\begin{tabular}{l|llllll}
\hline\noalign{\smallskip}
Sh classes & \{0,1\} & \{0,2\} & \{1,2\} & \{0\} & \{1\} & \{2\}\\
Rj classes & \{2\}& \{1\} & \{0\} & \{1,2\} & \{0,2\} & \{0,1\}\\
\hline\noalign{\smallskip}
\% of Rj classes & 33\% & 33\% & 33\% & 66\% & 66\% & 66\%\\
\hline\noalign{\smallskip}
1Vs (Linear) & 0.5 & 0.37 & 0.49 & 0 & 0 & 0 \\
\hline\noalign{\smallskip}
WSVM (RBF) & 0.81 & 0.8 & 0.79 & - & - & -\\
\hline\noalign{\smallskip}
PISVM (RBF) & 0.94 & 0.83 & 0.8 & 0.5 & 0.5 & 0.5\\
\hline\noalign{\smallskip}
Ours &  \textbf{0.95} & \textbf{0.96} & \textbf{0.97} & \textbf{0.98} & \textbf{0.98} & \textbf{0.96}\\
\hline
\end{tabular}
\end{center}
\setlength{\tabcolsep}{1.4pt}
\vspace*{-5mm}
\end{table}

\subsubsection{Experiments on real datasets}
For this step, we first evaluate our rejection algorithm on datasets under Label-Shift and open set classes.
We modify the set of classes for each experiment in order to test different proportions of \emph{rejected class}. 
We use  USPS~(U), MNIST~(M) and SVHN~(S) benchmarks.
All the benchmarks contain 10 classes.
USPS images have single channel and a size of $16\times 16$ pixels, MNIST images have single channel and a size of $28 \times 28$ pixels while SVHN images have 3-color channels and a size of $32\times 32$ pixels.

As a first experiment, we sample our source and target datasets from the same benchmark {i.e.} USPS $\rightarrow$ USPS, MNIST $\rightarrow$ MNIST and SVHN $\rightarrow$ MNIST.
We choose different samples for each domain and modify the set of shared and rejected classes.
Then, we present challenging cases with increasing Covariate- Shift as source and target samples are from different benchmarks as shown in Table~\ref{table:f1realmnist}.
For each benchmark, we resize the images to $32 \times 32$ pixels  and split source samples into training and test sets. We extract feature embeddings using the following process :
\begin{enumerate*}[label=\arabic*)]
\item We train a Neural Network (as suggested in ~\cite{Haeusser_2017_ICCV}) on the training set of source domain, 
\item We randomly sample 200 images (except for USPS 72 images instead) for each class from test set of source and target domains, and
\item We extract image embeddings of chosen samples from the last Fc layer (128 units) of the trained model. 
\end{enumerate*}

We compare our Rejection algorithm to the $1$-Vs Machine~\cite{article} , PISVM~\cite{10.1007/978-3-319-10578-9_26} and WSVM~\cite{6809169}\footnote{ \url{https://github.com/ljain2/libsvm-openset}} which are based on SVM and require a threshold to provide a decision.
For tasks with a single rejected class, we get results similar to PISVM and WSVM when noise is small (Table~\ref{table:f1synth1}) and outperfom all methods when noise increases (Table~\ref{table:f1synth5}).
These results prove that we are more robust to ambiguous dataset. For tasks with multiple rejected classes, WSVM is not suitable to this case and PISVM and 1Vs performs poorly compared to our approach.
In fact, these approaches strongly depend on openness measure~\cite{10.1007/978-3-319-10578-9_26,6809169}.

As for the case with small noise, we obtain similar results for DA tasks with Label-Shift only as shown in Table~\ref{table:f1realmnist} while we outperform state-of-art methods for DA tasks combining target and covariate shifts (Table~\ref{table:Rejection MNIST USPS}) except for last task where WSVM slightly exceeds our method.
This confirms the ability of our approach to address challenging shifts.
In addition, our proposed approach for the rejection step is based on OT which provides a framework consistent with the Label-Shift step.

\begin{table}[h]
\vspace*{-5mm}
\setlength{\tabcolsep}{4pt}
\begin{center}
\caption{F1-score of Rejection algorithms applied to target samples of MNIST benchmark}
\vspace*{-5mm}
\label{table:f1realmnist}
\begin{tabular}{l|llllll}
\hline\noalign{\smallskip}
{\small Sh classes} & \small{\{0,2,4\}} & \small{ \{6,8\}} & \small{ \{1,3,5\}} & \small{ \{7,9\}} & \small{ \{0,1,2,3,4\}}\\
{\small Rj classes} & \small{ \{6,8\} }& \small{ \{0,2,4\}} & \small{ \{7,9\} }& \small{ \{1,3,5\}} & \small{ \{5,6,7,8,9\} }\\
\hline\noalign{\smallskip}
{\small \% of Rj classes } & \small{ 40\% }& \small{ 60\% } & \small{ 40\% } & \small{ 60\% } & \small{ 50\%} \\
\hline\noalign{\smallskip}
{\small 1Vs (Linear)}& \small{0.65 $\pm$ 0.01} & \small{0} & \small{0.74 $\pm$ 0.01} & \small{0.29 $\pm$ 0.04} & {\small 0.61} \\
\hline\noalign{\smallskip}
\small{WSVM (RBF)}& \small{0.97 $\pm$ 0.02} & \small{0.95 $\pm$ 0.0 }& \small{\textbf{0.98} $\pm$ 0.01} & \small{0.76 $\pm$ 0.2 }& \small{0.96 $\pm$ 0.01}\\
\hline\noalign{\smallskip}
\small{PISVM (RBF)}& \small{\textbf{0.98} $\pm$ 0.01} & \small{0.96 $\pm$ 0.02 }& \small{\textbf{0.98} $\pm$ 0.01 }& \small{0.80 $\pm$ 0.16} & \small{\textbf{0.97} $\pm$ 0.01}\\
\hline\noalign{\smallskip}
\small{Ours} & \small{\textbf{0.98} $\pm$ 0.01} & \small{\textbf{0.99} $\pm$ 0.01 }& \small{\textbf{0.98} $\pm$ 0.01} & \small{\textbf{0.97} $\pm$ 0.01 }& \small{0.93 $\pm$ 0.02}\\
\hline
\end{tabular}
\end{center}
\setlength{\tabcolsep}{1.4pt}
\vspace*{-10mm}
\end{table}

\begin{table}[h]
\setlength{\tabcolsep}{4pt}
\begin{center}
\caption{F1-score of Rejection algorithms applied to target samples where source domain: MNIST and target domain: USPS}
\label{table:Rejection MNIST USPS}
\vspace*{-5mm}
\begin{small}
\begin{tabular}{l|llllll}
\hline\noalign{\smallskip}
Sh classes & \{0,2,4\} & \{6,8\} &\{1,3,5\} & \{7,9\} & \{0,1,2,3,4\}\\
Rj classes & \{6,8\}& \{0,2,4\} & \{7,9\} & \{1,3,5\} & \{5,6,7,8,9\}\\
\hline\noalign{\smallskip}
\% of Rj classes & 40\% & 60\% & 40\% & 60\% & 50\%\\
\hline\noalign{\smallskip}
1Vs (Linear)& 0.57 $\pm$ 0.04 & 0 & 0.62 $\pm$ 0.05 & 0.27 $\pm$ 0.06 & 0.53 $\pm$ 0.04 \\
\hline\noalign{\smallskip}
WSVM (RBF)& 0.82 $\pm$ 0.09 & 0.69 $\pm$ 0.07 & 0.86 $\pm$ 0.05 & 0.64 $\pm$ 0.06 & \textbf{0.79} $\pm$ 0.04 \\
\hline\noalign{\smallskip}
PISVM (RBF)& 0.82 $\pm$ 0.09 & 0.68 $\pm$ 0.06 & 0.86 $\pm$ 0.05 & 0.66 $\pm$ 0.06 & 0.77 $\pm$ 0.04 \\
\hline\noalign{\smallskip}
Ours & \textbf{0.9} $\pm$ 0.02 & \textbf{0.83} $\pm$ 0.03 & \textbf{0.87} $\pm$ 0.05 & \textbf{0.92} $\pm$ 0.02 & 0.74 $\pm$ 0.06\\
\hline
\end{tabular}
\end{small}
\end{center}
\setlength{\tabcolsep}{1.4pt}
\vspace*{-5mm}
\end{table}

\subsection{Label-Shift}

We sample unbalanced source datasets and reversely unbalanced target datasets for both MNIST and SVHN benchmarks in order to create significant Label-Shift as shown in Fig.~\ref{fig:artificial label shift} in Appendix~\ref{ap:artificial label shift}. USPS benchmark is too small (2007 samples for test) and is already unbalanced. Therefore we use all USPS samples for the experiments M$\rightarrow$U and U$\rightarrow$M.

We create 5 tasks by increasing Covariate-Shift to evaluate the robustness of our algorithm.
We compare our approach to JDOT \cite{courty2017joint} and JCPOT \cite{ievgen_2019} which predicts  target label in two different ways (label propagation JCPOT-LP and JCPOT-PT).
We used the public code given by the authors for JDOT\footnote{Code available at \url{https://github.com/rflamary/JDOT}} and JCPOT\footnote{Code available at \url{https://github.com/ievred/JCPOT}}.
Note that JCPOT is applied to multi-source samples. Consequently,we split \{$X^s,Y^s$\} into $N$ sources with random class proportions and chose $N$ which gives the best results (N=5). We present the results on 5 trials.
We set $\eta=0.001$ for all experiments with the Label-Shift algorithm.
JCPOT uses a grid search to get its optimal $\eta$.

For synthetic dataset, JCPOT and our Label-Shift method give similar results (Table~\ref{table:label shift synthetic}).
However, for real datasets as shown in Table~\ref{table:lable shift digits}, we widely outperform other state-of-the-art DA methods especially for DA tasks that present covariate shift in addition to the Label-Shift.
These results prove that our approach is more robust to high-dimensional dataset as well as to distributions with combined label and covariate shifts.

\begin{table}[h]
\setlength{\tabcolsep}{4pt}
\begin{center}
\caption{Recorded F1-score for  Label-Shift  algorithms  applied  to synthetic datasets.}
\label{table:label shift synthetic}
\vspace*{-1mm}
\begin{tabular}{l|llll}
\hline\noalign{\smallskip}
Setting & JDOT & JCPOT-LP(5) & JCPOT-PT(5) & Ours\\
\hline\noalign{\smallskip}
Noise = 0.5 & 0.5 & 0.997 &	0.99 &	0.997\\
\hline\noalign{\smallskip}
Noise = 0.75 &  0.45 & 0.98	& 0.94 & 0.98\\
\hline
\end{tabular}
\end{center}
\setlength{\tabcolsep}{1.4pt}
\vspace*{-10mm}
\end{table}

\begin{table}[h]
\setlength{\tabcolsep}{4pt}
\begin{center}
\caption{F1-score  of Label-Shift algorithms on digits classification tasks.} 
\label{table:lable shift digits}
\vspace*{-5mm}
\begin{small}
\begin{tabular}{l|lllll}
\hline\noalign{\smallskip}
Methods & M$\rightarrow$M & S$\rightarrow$S & M$\rightarrow$U & U$\rightarrow$M & S$\rightarrow$M\\
\hline\noalign{\smallskip}
JDOT & 0.52 $\pm$ 0.04 & 0.53 $\pm$ 0.01 & 0.64 $\pm$ 0.01 & 0.87 $\pm$ 0.02 & 0.43 $\pm$ 0.01\\
\hline\noalign{\smallskip}
JCPOT-LP(5) & \textbf{0.98} $\pm$ 0.002 & 0.37 $\pm$ 0.43 & 0.56 $\pm$ 0.0.026 & 0.89 $\pm$ 0.01 & 0.21 $\pm$ 0.237 \\
\hline\noalign{\smallskip}
JCPOT-PT(5) & 0.96 $\pm$ 0.004 & 0.81 $\pm$ 0.045 & 0.40 $\pm$ 0.327 & 0.86 $\pm$ 0.013 & 0.46 $\pm$ 0.222\\
\hline\noalign{\smallskip}
Ours & \textbf{0.98} $\pm$ 0.001& \textbf{0.92} $\pm$ 0.006 & \textbf{0.76} $\pm$ 0.019 & \textbf{0.92} $\pm$ 0.006 & \textbf{0.65} $\pm$ 0.017\\
\hline
\end{tabular}
\end{small}
\end{center}
\setlength{\tabcolsep}{1.4pt}
\vspace{-10mm}
\end{table}

\subsection{Full 2-step approach: Rejection and Label-Shift}

The same shared and rejected classes from the rejection experiments tasks have been chosen.
We also create significant Label-Shift as done for Label-Shift experiments (Unbalanced and Reversely-unbalanced class proportions) for synthetic datasets as well as for MNIST and SVHN real benchmarks.
Nevertheless, we keep the initial class proportions of USPS due to the size constraint of the database. This time, we implement a jointly 2-step. Namely, we plug the obtained target marginal in the Label-Shift step after discarding rejected samples. We apply Algorithm~\ref{algo_RV} to rejection step to get optimal hyperparameters ($\eta$,$\alpha$) and keep the same $\eta$  for Label-shift step. We obtained $\eta=0.001$ and $\alpha=1$.

In Table~\ref{table:combination synthetic}, we show results for synthetic data generated with different noises.
When noise increases, i.e., boundary decision between classes is ambiguous, the performance is affected. 
Table~\ref{table:combination real} presents F1-score over 10 runs of our 2-step approach applied to real datasets. For DA tasks with only Label-Shift (M$\rightarrow$M and S$\rightarrow$S), F1-score is high. However it drops when we address both Covariate and Label-Shift (M$\rightarrow$U, U$\rightarrow$M and S$\rightarrow$M).
In fact, previous results for each step (Tables~\ref{table:Rejection MNIST USPS} and~\ref{table:lable shift digits}) have shown that performance was affected by Covariate-Shift.
The final result of our 2-step approach is linked to the performance of each separate step. We present an illustration of the full algorithm in Fig.~\ref{fig: full algo} in Appendix~\ref{ap: full algo}.

\begin{table}[h]
\setlength{\tabcolsep}{4pt}
\begin{center}
\caption{F1-score across target samples of combined our 2-step approach applied to synthetic data, $\eta$ $=$ 0.001, $\alpha$=1}
\label{table:combination synthetic}
\begin{tabular}{l|lll}
\hline\noalign{\smallskip}
Sh classes & \{0,1\} & \{0,2\} & \{1,2\}\\
Rj classes & \{2\}& \{1\} & \{0\}\\
\hline\noalign{\smallskip}
Noise = 0.5 &  1 & 0.99 & 0.99\\
\hline\noalign{\smallskip}
Noise = 0.75 &  0.93 & 0.87 & 0.85\\
\hline
\end{tabular}
\end{center}
\setlength{\tabcolsep}{1.4pt}
\vspace*{-10mm}
\end{table}

\begin{table}[h]
\setlength{\tabcolsep}{4pt}
\begin{center}
\caption{F1-score across target samples of our combined 2-step approach applied to real datasets features, $\eta$=0.001, $\alpha$=1 }
\label{table:combination real}
\vspace*{-5mm}
\begin{small}
\begin{tabular}{l|llllll}
\hline\noalign{\smallskip}
Benchmarks & M$\rightarrow$M & S$\rightarrow$S & M$\rightarrow$U & U$\rightarrow$M & S$\rightarrow$M\\
\hline\noalign{\smallskip}
Sh \{0,2,4\} & 0.93 $\pm$ 0.005 & 0.91 $\pm$ 0.008 & 0.65 $\pm$ 0.011 &  0.59 $\pm$ 0.014  & 0.66 $\pm$ 0.011\\
Rj \{6,8\}\\
\hline\noalign{\smallskip}
Sh \{6,8\} & 0.95 $\pm$ 0.006 & 0.89 $\pm$ 0.012 & 0.82 $\pm$ 0.013 & 0.61 $\pm$ 0.01 & 0.53 $\pm$ 0.014\\
Rj \{0,2,4\} & \\
\hline\noalign{\smallskip}
Sh \{1,3,5\} & 0.93 $\pm$ 0.009 & 0.86 $\pm$ 0.01 & 0.76 $\pm$ 0.02 & 0.58 $\pm$ 0.011 & 0.74 $\pm$ 0.018 \\
Rj \{7,9\} & \\
\hline\noalign{\smallskip}
Sh \{7,9\} & 0.97 $\pm$ 0.009 & 0.90 $\pm$ 0.011 & 0.75 $\pm$ 0.008 & 0.52 $\pm$ 0.007 & 0.65 $\pm$  0.021 \\
Rj \{1,3,5\} & \\
\hline\noalign{\smallskip}
Sh \{0,1,2,3,4\} & 0.91 $\pm$ 0.01 & 0.82 $\pm$ 0.007 & 0.73 $\pm$ 0.013  & 0.74 $\pm$ 0.01  & 0.68 $\pm$ 0.01 \\
Rj \{5,6,7,8,9\} & \\
\hline
\end{tabular}
\end{small}
\end{center}
\setlength{\tabcolsep}{1.4pt}
\vspace*{-5mm}
\end{table}

\section{Conclusion}
In this paper, we proposed an optimal transport framework to solve open set DA. It is composed of two steps solving 
Rejection and Label-shift adaptation problems. The main idea was to learn the transportation plans together with the marginal distributions. 
Notably, experimental evaluations showed that applying our algorithms to various datasets lead to consistent outperforming results over the state-of-the-art.  
We plan to extend the framework to learn deep networks for open set domain adaptation. 

\section*{Acknowledgements} This work was supported by the National Research Fund, Luxembourg (FNR) and the OATMIL ANR-17-CE23-0012 Project of the French National Research Agency (ANR).

\FloatBarrier

%
%

\ifbibfile
\bibliographystyle{splncs04}
\bibliography{egbib}
\else


\fi

\appendix
\newpage
\section{Appendix}

\subsection{Proof of Lemma~\ref{lemma_rejection}}
\label{ap:lemma_rejection}

Define the dual Lagrangian function 
\begin{align*}
&\mathscr{L}_{\rj}(\gamma, \mu^t, \lambda, \beta, \vartheta, \theta)\\
&= \inr{\zeta, \gamma}_F - \eta H(\gamma) + \inr{\lambda, \gamma \mathbf 1_{n_t} -\mu^s} + \inr{\beta, \gamma^\top \mathbf 1_{n_s} -\mu^t} - \inr{\vartheta,\mu^t} + \theta (\norm{\mu^t}_1 -1)\\
&= \inr{\zeta, \gamma}_F - \eta H(\gamma) + \inr{\lambda, \gamma \mathbf 1_{n_t}} + \inr{\beta, \gamma^\top \mathbf 1_{n_s}} 
- \inr{\beta,\mu^t} + \theta \norm{\mu^t}_1 -  \inr{\vartheta,\mu^t}
- \inr{\lambda, \mu^s}  - \theta
\end{align*}
equivalently
\begin{align*}
 \mathscr{L}_{\rj}(\gamma, \mu^t, \lambda, \beta, \theta)  = E_{\rj}(\gamma) + F_{\rj}(\mu^t) + G_{\rj}(\lambda, \theta),
\end{align*}
where 
\begin{align*}
E_{\rj}(\gamma) = \inr{\zeta, \gamma}_F - \eta H(\gamma) + \inr{\lambda, \gamma \mathbf 1_{n_t}} + \inr{\beta, \gamma^\top \mathbf 1_{n_s}}, 
\end{align*}
\begin{equation*}
F_{\rj}(\mu^t) = - \inr{\beta,\mu^t} -  \inr{\vartheta,\mu^t}+ \theta \norm{\mu^t}_1, \text{ and }  G_{\rj}(\lambda, \theta) = - \inr{\lambda, \mu^s}  - \theta.
\end{equation*}
We have 
\begin{align*}
\frac{\partial \mathscr{L}_{\rj}(\gamma, \mu^t, \lambda, \beta, \vartheta,\theta)}{\partial \gamma_{ij}} = \frac{\partial E_{\rj}(\gamma)}{\partial \gamma_{ij}}  = C_{ij} + \eta(\log \gamma_{ij} + 1) + \lambda_j + \beta_j,
\end{align*}
and 
\begin{align*}
\frac{\partial \mathscr{L}_{\rj}(\gamma, \mu^t, \lambda, \beta, \vartheta,\theta)}{\partial \mu^t_j} = \frac{\partial F_{\rj}(\mu^t)}{\partial \mu^t_j} = -\beta_j -\vartheta_j + \theta.
\end{align*}
Then the couple $(\gamma^\star_{\rj}, {\mu^t}^\star)$ optimum of the dual Lagrangian function $\mathscr{L}_{\rj}(\gamma, \mu^t, \lambda, \beta, \theta)$ satisfies the following 
\begin{align*}
\begin{cases}
\frac{\partial \mathscr{L}_{\rj}(\gamma^\star_{\rj}, (\mu^t)^\star, \lambda, \beta,\vartheta, \theta)}{\partial \gamma^\star_{ij}} = 0\\
\frac{\partial \mathscr{L}_{\rj}(\gamma, (\mu^t)^\star, \lambda, \beta, \vartheta,\theta)}{\partial {\mu^t}^\star_{ij}} = 0
\end{cases}
\equiv 
\begin{cases}
(\gamma^\star_{\rj})_{ij} = \exp\big(-\frac{C_{ij} + \lambda_i + \beta_j}{\eta} - 1\big),\\
\theta - \beta_j = 0, 
\end{cases}
\end{align*}
for all $i=1, \ldots, n_s$ and $j =1, \ldots, n_t.$ Now, plugging this solution in the Lagrangian function 
we get 
\begin{align*}
\mathscr{L}_{\rj}(\gamma^\star_{\rj}, {\mu^t}^\star, \lambda, \beta, \vartheta, \theta)
&= \sumins\sumint C_{ij} \exp\big(-\frac{C_{ij} + \lambda_i + \beta_j}{\eta} - 1\big)\\
& \qquad + \eta \sumins\sumint (-\frac{C_{ij} + \lambda_i + \beta_j}{\eta} - 1) \exp\big(-\frac{C_{ij} + \lambda_i + \beta_j}{\eta} - 1\big)\\
& \qquad+ \sumins \lambda_i \sumint \exp\big(-\frac{C_{ij} + \lambda_i + \beta_j}{\eta} - 1\big)\\
& \qquad+ \sumint \beta_j \sumins \exp\big(-\frac{C_{ij} + \lambda_i + \beta_j}{\eta} - 1\big)\\
& \qquad- \inr{\beta,{\mu^t}^\star} - \inr{\vartheta,{\mu^t}^\star} + \theta \norm{{\mu^t}^\star}_1, - \inr{\lambda, \mu^s}  - \theta.
\end{align*}
Note that $- \inr{\beta,{\mu^t}^\star} - \inr{\vartheta,{\mu^t}^\star} + \theta \norm{{\mu^t}^\star}_1 = \inr{-\beta - \vartheta + \theta \mathbf 1_{n_t}, {\mu^t}^\star}$, hence taking into account the constraint $\theta - \beta_j -\vartheta_j= 0,$ for all $j =1, \ldots, n_t,$ it entails that $- \inr{\beta + \vartheta,{\mu^t}^\star} + \theta \norm{{\mu^t}^\star}_1 = 0.$
Hence 
\begin{align*}
&\mathscr{L}_{\rj}(\gamma^\star_{\rj}, {\mu^t}^\star, \lambda, \beta, \vartheta, \theta) = -\eta \sumins\sumint \exp\big(-\frac{C_{ij} + \lambda_i + \beta_j}{\eta} - 1\big) - \inr{\lambda, \mu^s}  - \theta,
\end{align*}
subject to $\theta - \beta_j - \vartheta_j= 0$ for all $j =1, \ldots, n_t.$ Setting the following variable change $f = -\frac\lambda\eta - \frac 12 \mathbf 1_{n_s}$ and $g = -\frac\beta\eta - \frac 12 \mathbf 1_{n_t}$ we get 
\begin{align*}
\mathscr{L}_{\rj}(\gamma^\star_{\rj}, {\mu^t}^\star, \lambda, \beta, \vartheta, \theta)
&= -\eta \sumins\sumint \exp\big(-\frac{C_{ij}}{\eta} + f_i + g_j\big) + \eta\inr{(f + \frac 12 \mathbf 1_{n_s}), \mu^s}  - \theta\\
& = -\eta \sumins\sumint \exp\big(-\frac{C_{ij}}{\eta} + f_i + g_j\big) + \eta \inr{f, \mu^s} + \eta\frac 12 - \theta\\
& = -\eta \mathbf 1_{n_s}^\top B(f, g) \mathbf 1_{n_t} + \eta \inr{f, \mu^s} + \eta\frac 12 - \theta.
\end{align*}
Then 
\begin{align*}
\mathscr{L}_{\rj}(\gamma^\star_{\rj}, {\mu^t}^\star, \lambda, \beta, \vartheta, \theta) = - \eta\big\{\mathbf 1_{n_s}^\top B(f, g) \mathbf 1_{n_t} -\inr{f, \mu^s} - \frac 12 + \frac\theta\eta\big\},
\end{align*}
subject to $\theta + \eta(g_j + \frac 12) = 0$. Putting $\kappa = \frac\theta\eta - \frac 12$, then $\theta = \eta(\kappa + \frac 12)$. This gives
\begin{align*}
\mathscr{L}_{\rj}(\gamma^\star_{\rj}, {\mu^t}^\star, \lambda, \beta, \vartheta, \theta) \equiv \mathscr{L}_{\rj}(\gamma^\star_{\rj}, {\mu^t}^\star, \lambda, \beta, \kappa) = - \eta\big\{\mathbf 1_{n_s}^\top B(f, g) \mathbf 1_{n_t} -\inr{f, \mu^s} + \kappa\big\},
\end{align*}
subject to $g_j + \kappa + 1 = 0$, for all $j =1, \ldots, n_t.$ 
We remark that
\begin{equation*}
\mathbf 1_{n_s}^\top B(f, g) \mathbf 1_{n_t} = \sumins\sumint e^{f_i - \kappa} K_{ij} e^{g_j + \kappa} =\mathbf 1_{n_s}^\top B(f -\kappa\mathbf 1_{n_s}, g + \kappa\mathbf 1_{n_t})\mathbf 1_{n_t},
\end{equation*}
then using a variable change $\tilde{f} = f -\kappa\mathbf 1_{n_s}$ and $\tilde{g} = g +\kappa\mathbf 1_{n_t}$ we get 
\begin{align*}
(f^\star_{\rj}, g^\star_{\rj}) 
= \argmin_{\substack{\tilde{f} \in \R^{n_s}, \tilde{g} \in \R^{n_t},\\ \tilde{g}_j + 1 = 0, \forall j=1, \ldots, n_t}} \{\mathbf 1_{n_s}^\top B(\tilde{f}, \tilde{g}) \mathbf 1_{n_t} -\inr{\tilde{f}, \mu^s}\}.
\end{align*}
Therefore
\begin{equation*}
(f^\star_{\rj}, g^\star_{\rj}) 
= \argmin_{\substack{f \in \R^{n_s}, g \in \R^{n_t}}} \{\mathbf 1_{n_s}^\top B(f, g) \mathbf 1_{n_t} -\inr{f, \mu^s} + \chi_{-\mathbf 1_{n_t}} (g)\}.
\end{equation*}

\subsection{Proof of Lemma~\ref{lem_kkt_rejection}}

Setting 
\begin{equation*}
\Psi(f, g) = \mathbf 1_{n_s}^\top B(f, g) \mathbf 1_{n_t} -\inr{f, \mu^s} + \chi_{-\mathbf 1_{n_t}} (g),
\end{equation*}
Writting the KKT optimlaity condition for the above problem leads to the following: we have $f \mapsto \Psi(f, g)$ is differentiable, hence we can calculate a gradient with respect to $f$. However $g \mapsto \Psi(f, g$ is not differentiable, then we just  calculate a subdifferentiale as follows: 
\begin{align*}
\nabla \Psi(f, g) = \Big\{e^{f_i} \sumint K_{ij}e^{g_j} - \mu^s_i\Big\}_{1 \leq i \leq n_s} \in \R^{n_S},
\end{align*}
and 
\begin{equation*}
\partial_{g}(\Psi(f, g)) = \Big\{e^{g_j}\sumins K_{ij} e^{f_i} + \partial_{}(\chi_{-\mathbf 1_{n_t}} (g))\Big\}_{1 \leq j \leq n_t},
\end{equation*}
where $\partial_{g}(\chi_{-\mathbf 1_{n_t}} (g))$ is the subdifferential of the indicator function $\chi_{-\mathbf 1_{n_t}}$ at $g$ is known as the normal cone, namely
\begin{align*}
\partial_{}(\chi_{-\mathbf 1_{n_t}} (g)) &= \{\bu \in \R^{n_t}| \bu^\top g \geq -\bu^\top \mathbf 1_{n_t}\}\\
&= \Big\{\bu \in \R^{n_t}| \sumint u_jg_j \geq -\sumint u_j\Big\}.
\end{align*}
Therefore, KKT optimality conditions give 
\begin{equation*}
e^{f^\star_{\rj}} = \frac{\mu}{K e^{g^\star_{\rj}}} \text{ and } 
-e^{g^\star_{\rj}} \cdot K^\top e^{f^\star_{\rj}} \in  \partial_{}(\chi_{-\mathbf 1_{n_t}} (g^\star_{\rj})),
\end{equation*}
(the division $/$ and the multiplcation $\cdot$ between vectors have to be understood elementwise).
So 
\begin{equation*}
e^{(f^\star_{\rj})_i} = \frac{\mu^s_i}{\sumint K_{ij}e^{(g^\star_{\rj})_j}}\text{ and } 
-\sumint e^{(g^\star_{\rj})_j} K_{ij} e^{(f^\star_{\rj})_i} (g^\star_{\rj})_j \geq - (-\sumint e^{(g^\star_{\rj})_j} K_{ij} e^{(f^\star_{\rj})_i}),
\end{equation*}
equivalently
\begin{equation*}
e^{(f^\star_{\rj})_i} = \frac{\mu^s_i}{\sumint K_{ij}e^{(g^\star_{\rj})_j}}\text{ and } 
\sumint e^{(f^\star_{\rj})_i} K_{ij} e^{(g^\star_{\rj})_j} (1 + (g^\star_{\rj})_j) \leq 0.
\end{equation*}
for all $i=1, \ldots, n_s.$
 
\subsection{Proof of Lemma~\ref{lemma_shift}}
\label{ap:lemma_shift}

First, observe that $\Delta_C = \{\alpha \in \R^C_+: \mathbf 1_C^\top D\alpha = 1\}.$ Then the dual Lagrangian function is given by
\begin{align*}
&\mathscr{L}_{\ls}(\gamma, \nu, \lambda, \beta, \vartheta, \theta)\\
&= \inr{\zeta, \gamma}_F - \eta H(\gamma) + \inr{\lambda, \gamma \mathbf 1_{n_t} -D\nu} + \inr{\beta, \gamma^\top \mathbf 1_{n_s} -\mu^t} - \inr{\vartheta, \nu}+ \theta (\mathbf 1_C^\top D\nu -1)\\
&= \inr{\zeta, \gamma}_F - \eta H(\gamma) + \inr{\lambda, \gamma \mathbf 1_{n_t}} + \inr{\beta, \gamma^\top \mathbf 1_{n_s}} 
-\inr{\lambda, D\nu}- \inr{\beta,\mu^t} - \inr{\vartheta, \nu} + \theta \mathbf 1_C^\top D\nu
 - \theta,
\end{align*}
equivalently
\begin{align*}
 \mathscr{L}_{\ls}(\gamma, \mu^t, \lambda, \beta, \vartheta, \theta)  = E_{\ls}(\gamma) + F_{\ls}(\nu) + G_{\ls}(\lambda, \theta),
\end{align*}
where 
\begin{align*}
E_{\ls}(\gamma) = \inr{\zeta, \gamma}_F - \eta H(\gamma) + \inr{\lambda, \gamma \mathbf 1_{n_t}} + \inr{\beta, \gamma^\top \mathbf 1_{n_s}}, 
\end{align*}
\begin{equation*}
F_{\ls}(\nu) = -\inr{\lambda, D\nu} - \inr{\vartheta, \nu} + \theta \mathbf 1_C^\top D\nu, \text{ and }  G_{\ls}(\beta, \theta) = - \inr{\beta, \mu^s}  - \theta.
\end{equation*}
We have 
\begin{align*}
\frac{\partial \mathscr{L}_{\ls}(\gamma, \mu^t, \lambda, \beta, \vartheta,\theta)}{\partial \gamma_{ij}} = \frac{\partial E_{\ls}(\gamma)}{\partial \gamma_{ij}}  = C_{ij} + \eta(\log \gamma_{ij} + 1) + \lambda_j + \beta_j,
\end{align*}
and 
\begin{align*}
\frac{\partial \mathscr{L}_{\rj}(\gamma, \mu^t, \lambda, \beta, \vartheta,\theta)}{\partial \nu_c} = \frac{\partial F_{\ls}(\nu)}{\partial \nu_c} = -\sumins \lambda_id_{ic}+ \theta \sumins d_{ic} = \sumins(\theta - \lambda_i)d_{ic} -\vartheta_c
\end{align*}
Then the couple $(\gamma^\star_{\ls}, {\nu}^\star)$ optimum of the dual Lagrangian function $\mathscr{L}_{\ls}(\gamma, \nu, \lambda, \beta, \theta)$ satisfies the following 
\begin{align*}
\begin{cases}
\frac{\partial \mathscr{L}_{\ls}(\gamma^\star_{\ls}, \nu^\star, \lambda, \vartheta, \beta, \theta)}{\partial (\gamma^\star_{\ls})_{ij}} = 0\\
\frac{\partial \mathscr{L}_{\ls}(\gamma^\star_{\ls}, \nu^\star, \lambda, \beta, \vartheta,\theta)} {\partial {\nu}^\star_{c}} = 0
\end{cases}
\equiv 
\begin{cases}
(\gamma^\star_{\ls})_{ij} = \exp\big(-\frac{C_{ij} + \lambda_i + \beta_j}{\eta} - 1\big),\\
\sumins(\theta - \lambda_i)d_{ic} -\vartheta_c= 0, 
\end{cases}
\end{align*}

for all $i=1, \ldots, n_s$, $j =1, \ldots, n_t,$ and $c=1, \ldots, C.$ Now, plugging this solution in the Lagrangian function 
we get 
\begin{align*}
\mathscr{L}_{\ls}(\gamma^\star_{\ls}, {\nu}^\star, \lambda, \vartheta, \beta, \theta)
&= \sumins\sumint C_{ij} \exp\big(-\frac{C_{ij} + \lambda_i + \beta_j}{\eta} - 1\big)\\
& \quad + \eta \sumins\sumint (-\frac{C_{ij} + \lambda_i + \beta_j}{\eta} - 1) \exp\big(-\frac{C_{ij} + \lambda_i + \beta_j}{\eta} - 1\big)\\
& \quad+ \sumins \lambda_i \sumint \exp\big(-\frac{C_{ij} + \lambda_i + \beta_j}{\eta} - 1\big)\\
& \quad+ \sumint \beta_j \sumins \exp\big(-\frac{C_{ij} + \lambda_i + \beta_j}{\eta} - 1\big)\\
& \quad -\inr{\lambda, D\nu^\star} -\inr{\vartheta, \nu^\star}+ \theta \mathbf 1_C^\top D\nu^\star - \inr{\beta,\mu^t}  - \theta
\end{align*}
Observe that 
\begin{align*}
\theta \mathbf 1_C^\top D\nu^\star - \inr{\lambda, D\nu^\star}  - \inr{\vartheta, \nu^\star} &= \sum_{c=1}^C \sumins d_{ic}\nu^\star_c -  \sum_{c=1}^C \sumins d_{ic} \lambda_i d_{ic} \nu^\star_c - \sum_{c=1}^C \vartheta_c \nu^\star_c\\
&= \sum_{c=1}^C\Big(\sumins (\theta - \lambda)d_{ic} - \vartheta_c\Big)\nu^\star_c.
\end{align*}
Taking into account the constraint $\sumins(\theta - \lambda_i)d_{ic} - \vartheta_c= 0,$ for all $c =1, \ldots, C,$ it entails that $\theta \mathbf 1_C^\top D\nu^\star - \inr{\lambda, D\nu^\star} - \inr{\vartheta, \nu^\star}= 0.$
Hence 
\begin{align*}
&\mathscr{L}_{\ls}(\gamma^\star_{\ls}, {\nu}^\star, \lambda, \beta, \theta) = -\eta \sumins\sumint \exp\big(-\frac{C_{ij} + \lambda_i + \beta_j}{\eta} - 1\big) - \inr{\beta, \mu^t}  - \theta,
\end{align*}
subject to $\sumins(\theta - \lambda_i)d_{ic} - \vartheta_c= 0,$ for all $c =1, \ldots, C.$ Setting the following variable change $f = -\frac\lambda\eta - \frac 12 \mathbf 1_{n_s}$ and $g = -\frac\beta\eta - \frac 12 \mathbf 1_{n_t}$ we get 
\begin{align*}
\mathscr{L}_{\ls}(\gamma^\star_{\ls}, \nu^\star, \lambda, \beta, \vartheta, \theta)
&= -\eta \sumins\sumint \exp\big(-\frac{C_{ij}}{\eta} + f_i + g_j\big) + \eta\inr{(g + \frac 12 \mathbf 1_{n_t}), \mu^t}  - \theta\\
& = -\eta \sumins\sumint \exp\big(-\frac{C_{ij}}{\eta} + f_i + g_j\big) + \eta \inr{g, \mu^t} + \eta\frac 12 - \theta\\
& = -\eta \mathbf 1_{n_s}^\top B(f, g) \mathbf 1_{n_t} + \eta \inr{g, \mu^t} + \eta\frac 12 - \theta.
\end{align*}
Then 
\begin{align*}
\mathscr{L}_{\ls}(\gamma^\star_{\ls}, \nu^\star, \lambda, \beta, \vartheta, \theta) \equiv \mathscr{L}_{\ls}(\gamma^\star_{\ls}, {\nu}^\star, \lambda, \beta, \theta) = - \eta\big\{\mathbf 1_{n_s}^\top B(f, g) \mathbf 1_{n_t} -\inr{g, \mu^t} - \frac 12 + \frac\theta\eta\big\},
\end{align*}
subject to $\sumins(\theta + \eta(f_i + \frac 12))d_{ic} = 0$, for all $c=1, \ldots, C.$ Putting $\kappa = \frac\theta\eta - \frac 12$, then $\theta = \eta(\kappa + \frac 12)$. This gives 
\begin{align*}
\mathscr{L}_{\ls}(\gamma^\star_{\ls}, {\nu}^\star, \lambda, \beta, \kappa) = - \eta\big\{\mathbf 1_{n_s}^\top B(f, g) \mathbf 1_{n_t} -\inr{g, \mu^t} + \kappa\big\},
\end{align*}
subject to $\sumins(f_i + \kappa + 1)d_{ic} = 0$, for all $c =1, \ldots, C.$ 
We remark that
\begin{equation*}
\mathbf 1_{n_s}^\top B(f, g) \mathbf 1_{n_t} = \sumins\sumint e^{f_i - \kappa} K_{ij} e^{g_j + \kappa} =\mathbf 1_{n_s}^\top B(f -\kappa\mathbf 1_{n_s}, g + \kappa\mathbf 1_{n_t})\mathbf 1_{n_t}^\top,
\end{equation*}
then using a variable change $\tilde{f} = f + \kappa\mathbf 1_{n_s}$ and $\tilde{g}  = g -\kappa\mathbf 1_{n_t}$ we get 
\begin{align*}
(f^\star_{\ls}, g^\star_{\ls})
= \argmin_{\substack{\tilde{f}  \in \R^{n_s}, \tilde{g}  \in \R^{n_t},\\ \sumins(\tilde{f}_i + 1)d_{ic} = 0, \forall c=1, \ldots, C}} \{\mathbf 1_{n_s}^\top B(\tilde{f} , \tilde{g} ) \mathbf 1_{n_t} -\inr{\tilde{g}, \mu^t}\}.
\end{align*}
Finally 
\begin{align*}
(f^\star_{\ls}, g^\star_{\ls}) 
= \argmin_{\substack{f \in \R^{n_s}, g \in \R^{n_t},\\ \sumins (f_i  + 1)d_{ic} = 0, \forall c=1, \ldots, C}} \{\mathbf 1_{n_s}^\top B(f, g) \mathbf 1_{n_t} -\inr{g, \mu^t}\}
\end{align*}
that is 
\begin{align*}
(f^\star_{\ls}, g^\star_{\ls})  
= \argmin_{\substack{f \in \R^{n_s}, g \in \R^{n_t}}} \{\mathbf 1_{n_s}^\top B(f, g) \mathbf 1_{n_t} -\inr{g, \mu^t} + \chi_{\mathcal{F}}(f)\}.
\end{align*}

\begin{remark}
We omit the proof of Lemma~\ref{lem_kkt_shift} since it follows exactly the same lines as proof of Lemma~\ref{lem_kkt_rejection}.
\end{remark}

\clearpage

\subsection{Unbalancement trade-off}
\label{ap:artificial label shift}

\begin{figure}
\centering
\includegraphics[width=0.84\textwidth]{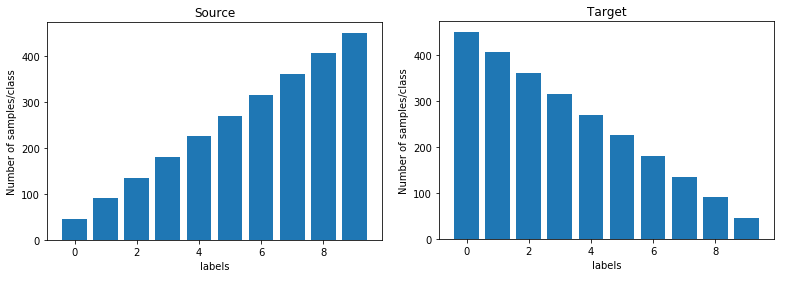}
\caption{Unbalanced source and reversely unbalanced target class proportions applied in label shift experiments as shown in Table~\ref{table:lable shift digits}}
\label{fig:artificial label shift}
\end{figure}


\subsection{Reverse validation details}
\label{ap:reverse validation}

\begin{algorithm}[htpb]
	\caption{Reverse Validation}
	\label{algo_RV}
	\textbf{require}: $\mathbf{\eta}$: list of suggested regularization terms; $\mathbf{\alpha}$: list of values to fix threshold, $X^s$: source samples; $\{X^t\}$: target samples; $n_s$: number of source samples; $n_t$ : number of target samples;\\
	\textbf{output}: $(\eta_i, \alpha_j)$: tuple of hyperparameters; 
	\begin{algorithmic}[1]
	    \State \textbf{initialize:}
		\State $k \gets 0, errors \gets [\,], hyperparameters \gets [\,];$ 
		\ForAll {$\eta_i$ in $\eta$}
		\ForAll {$\alpha_j$ in $\alpha$}
		\State $thresh \gets \alpha \frac{\eta}{n_s+n_t};$
		\State $\mu^t \gets \text{Rejection}(X^s, X^t,thresh)$;
		\State $X^t_{sc} \gets X^t[\mu^t > thresh];$
		\State $X^s_{new} \gets X^t_{sc}, X^t_{new} \gets X^s;$
		\State $\mu^t_{sc} \gets \text{Rejection}(X^s_{new}, X^t_{new},thresh);$
		\State $errors[k] \gets \frac{\#(\mu^t \leq thresh)}{n_s};$ 
		\State $hyperparameters[k] \gets (\eta_i,\alpha_j);$
		\State $k \gets k + 1;$
		\EndFor
		\EndFor
		\State \textbf{return:} $hyperparameters[\argmax(errors)];$
	\end{algorithmic} 
\end{algorithm} 

\clearpage
\subsection{Algorithm Illustration}
\label{ap: full algo}

\begin{figure}
\centering
\includegraphics[width=0.45\textwidth]{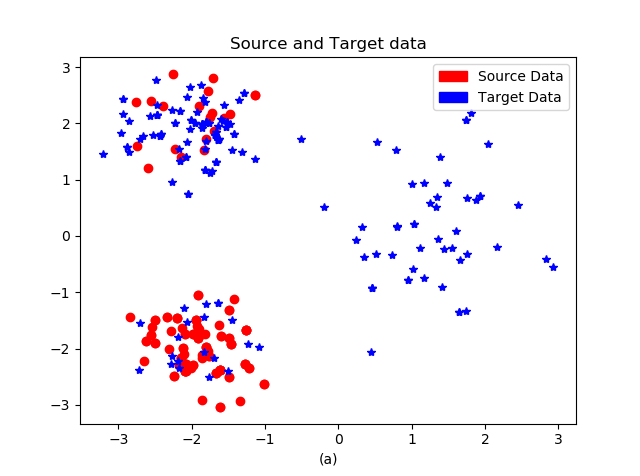}
\includegraphics[width=0.45\textwidth]{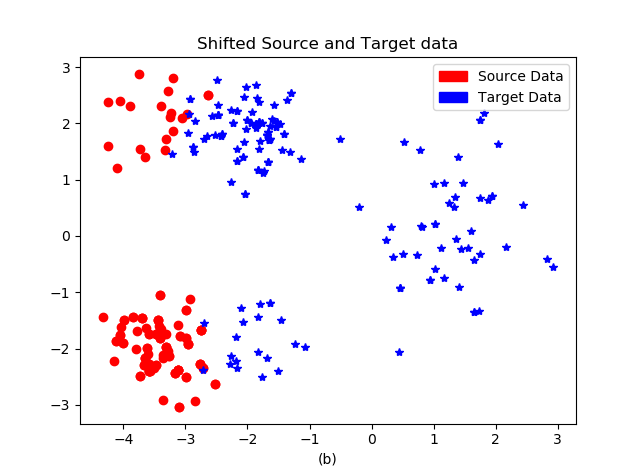}
\includegraphics[width=0.45\textwidth]{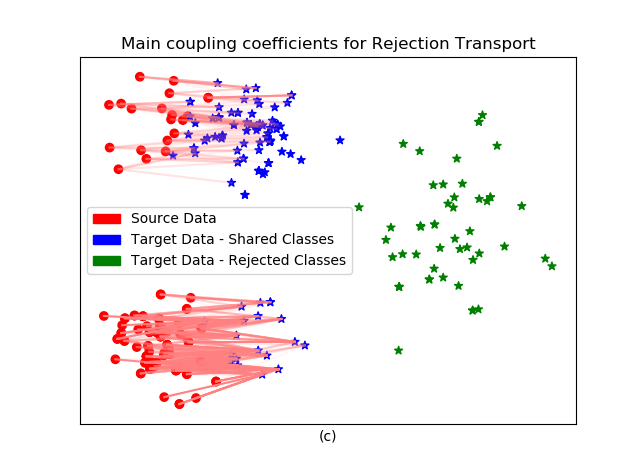}
\includegraphics[width=0.45\textwidth]{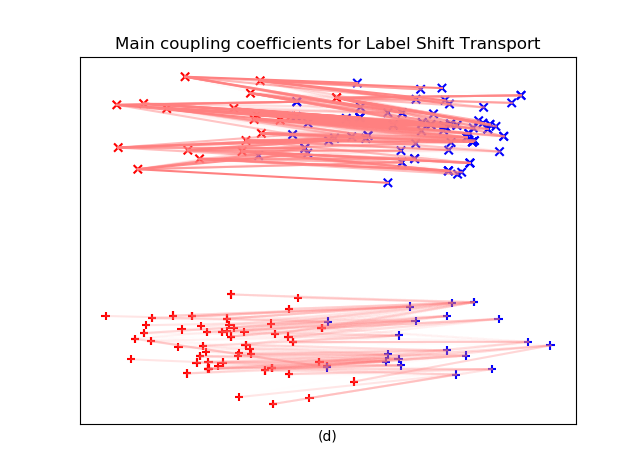}
\caption{Illustration of our 2-step approach for open set DA: 
(a)	Mixture of 2D Gaussian data where the classes \{0,1\} are the common classes between source and target domains and the class \{2\} is the rejected class in the target domain. The common classes' proportions for the source domain are $[0.25, 0.75]$ while for the target domain they are  chosen as $[0.75, 0.25]$; (b) Source data are slightly shifted from target data to ease visualization of mass transportation in figures (c) and (d); (c) Rejection step: Rejected points (in green) correspond to the points that receive a negligible amount of probability mass from source samples (d) Label-shift: mass transportation map obtained by label-shift algorithm.
}
\label{fig: full algo}
\end{figure}

\begin{figure}
\centering
\includegraphics[width=0.45\textwidth]{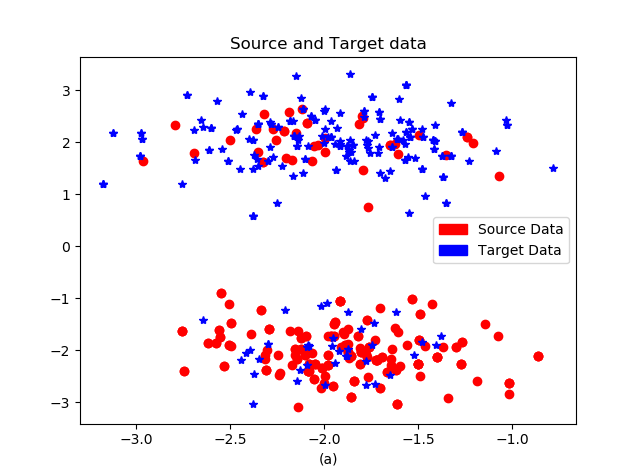}
\includegraphics[width=0.45\textwidth]{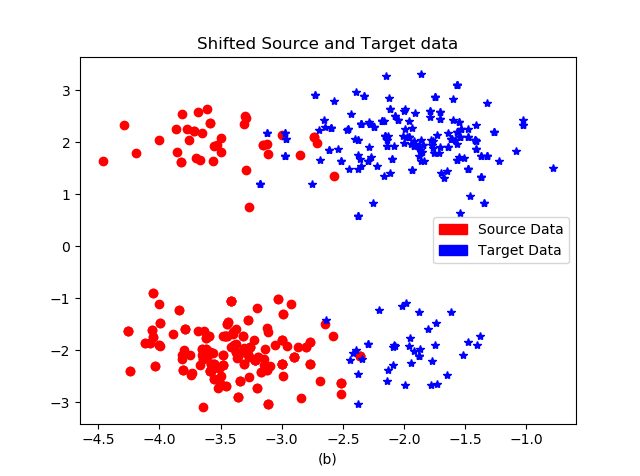}
\includegraphics[width=0.45\textwidth]{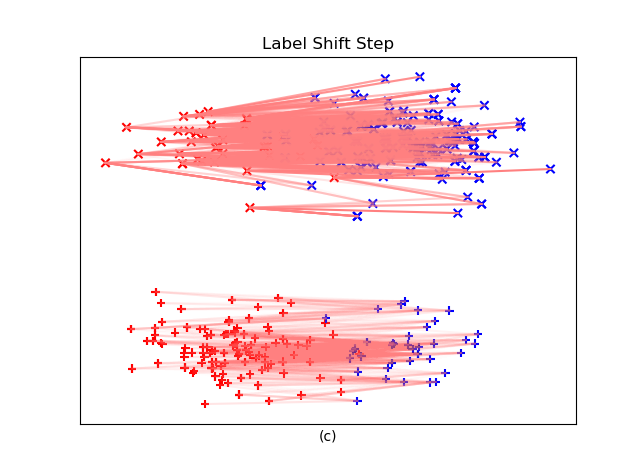}
\includegraphics[width=0.45\textwidth]{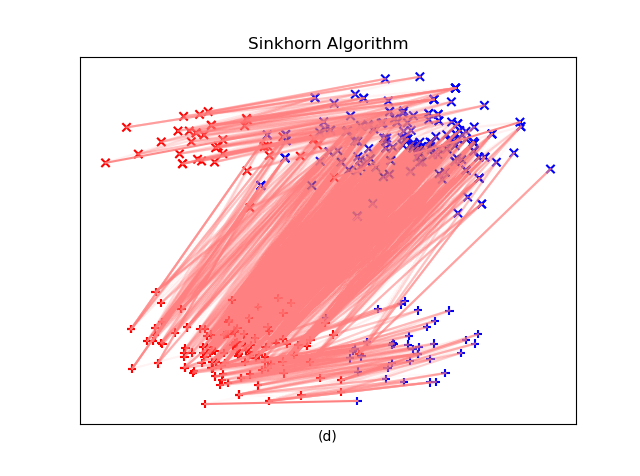}
\includegraphics[width=0.45\textwidth]{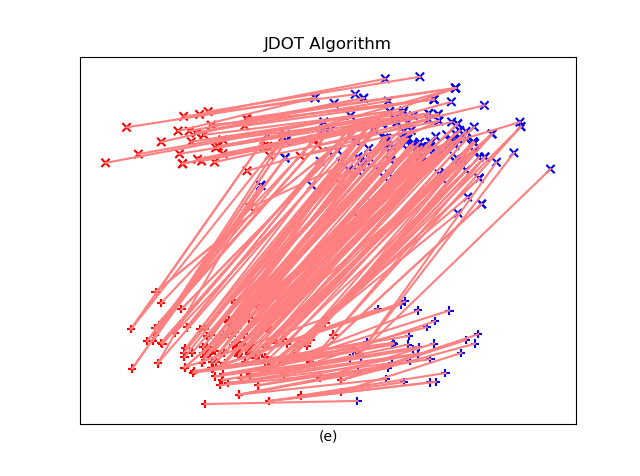}
\caption{Illustration of the Label-Shift on mixture of 2D Gaussian data with 2 classes such that the class proportions of the source domain are $[0.25, 0.75]$ while for the target domain are $[0.75, 0.25]$: (a) data; (b) Source data are slightly shifted from target data to ease visualization of mass transportation in figures (c), (d) and (e); (c) Our Label-Shift step;  (d) Unsupervised DA solution using
uniform source and target marginals.
Samples from one source class split their probability mass between target samples of both classes due to the existing label shift; (e) A similar behavior is observed for JDOT algorithm. 
}
\end{figure}

\end{document}